\documentclass[10pt,twocolumn,letterpaper]{article}

\usepackage[pagenumbers]{cvpr}

\usepackage{array}
\usepackage{algorithm}
\usepackage{algpseudocode}

\usepackage{caption}
\usepackage{graphicx}
\usepackage{multicol}
\usepackage{multirow}

\usepackage{float,wrapfig}

\usepackage{pifont}
\newcommand{\cmark}{\ding{51}}
\newcommand{\xmark}{\ding{55}}

\usepackage{amsmath}
\usepackage{dsfont}

\usepackage{makecell}
\definecolor{cvprblue}{rgb}{0.21,0.49,0.74}
\usepackage[pagebackref=false,breaklinks,colorlinks,allcolors=Emerald]{hyperref} 
\usepackage{marvosym}
\usepackage{caption}
\usepackage{fontawesome}

\title{
Zero-Shot 3D Visual Grounding from Vision-Language Models
}

\author{
    Rong Li$^\diamondsuit$\quad Shijie Li$^\triangle$\quad Lingdong Kong$^\heartsuit$\quad Xulei Yang$^\triangle$\quad Junwei Liang$^{\diamondsuit,\square,\textrm{\Letter}}$ 
    \\[1ex]
    $^\diamondsuit$HKUST(GZ)~~~
    $^\triangle$I$^2$R, A*STAR~~~
    $^\heartsuit$NUS~~~
    $^\square$CSE, HKUST
    \\[1ex]
    \faGithubAlt~\textbf{Project Page \& Code:} \href{https://seeground.github.io/}{\textsl{SeeGround.github.io}}
    \vspace{-0.7cm}
}

\begin{document}

\twocolumn[{
\renewcommand\twocolumn[1][]{#1}

\maketitle

\begin{center}
    \centering
    \includegraphics[width=\linewidth]{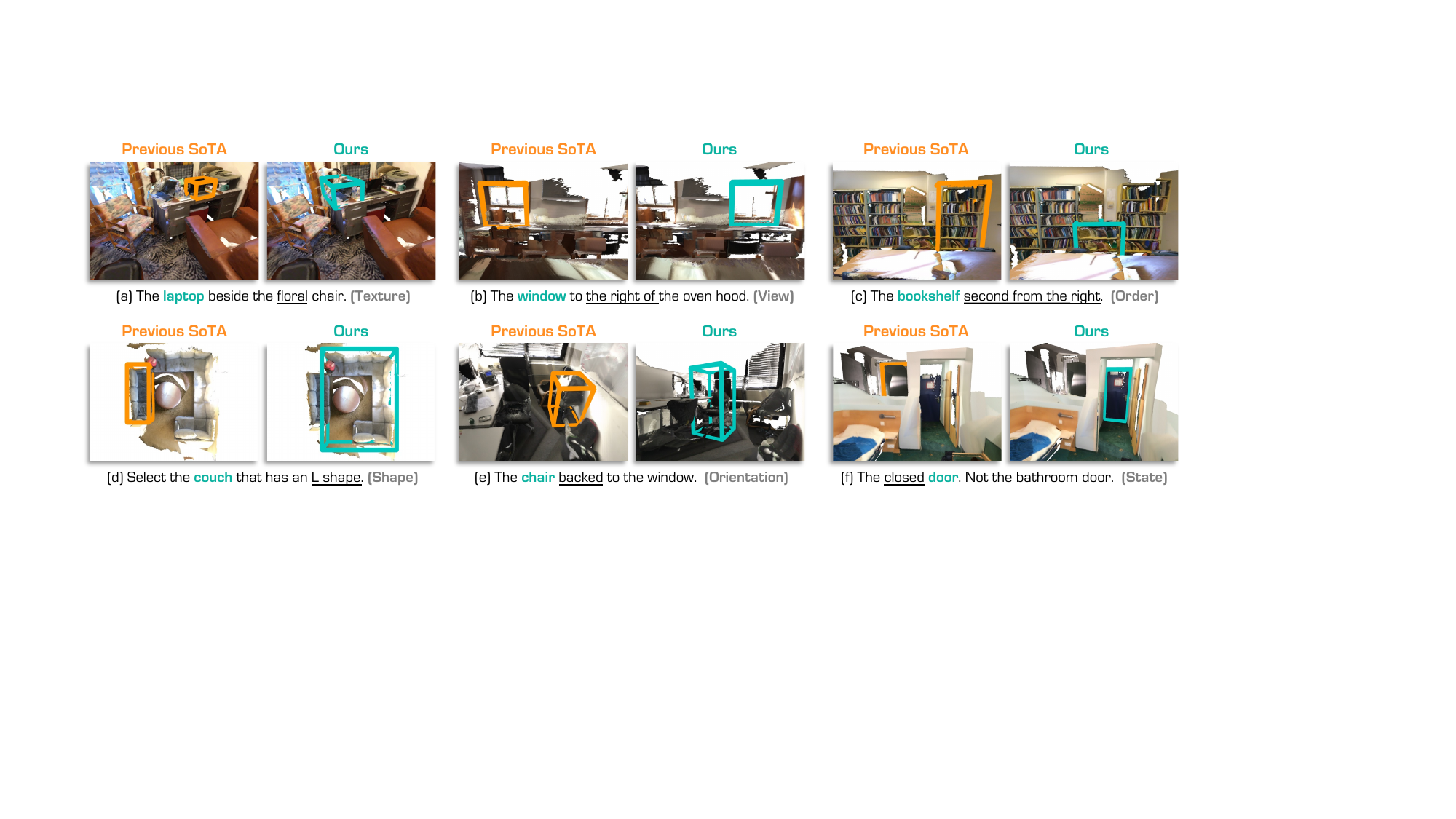}
    \vspace{-0.7cm}
    \captionof{figure}{Effectiveness of \textit{\textbf{\textcolor{BurntOrange}{See}\textcolor{Emerald}{Ground}}}:
    Unlike previous state-of-the-art methods, our approach aligns 2D visual cues -- such as \textit{\textbf{texture}}, \textit{\textbf{shape}}, \textit{\textbf{viewpoint}}, \textit{\textbf{spatial position}}, \textit{\textbf{orientation}}, \textit{\textbf{state}}, and \textit{\textbf{order}} -- with 3D spatial language to enable fine-grained scene comprehension. Specifically, our method: (a) \textit{\textbf{texture}}: detects the floral chair by leveraging distinctive color and texture patterns; (b) \textit{\textbf{shape}}: identifies the couch through its geometric shape; (c) \textit{\textbf{viewpoint}}: localizes the correct window by analyzing spatial relations and camera perspective; (d) \textit{\textbf{orientation}}: distinguishes the chair via directional alignment cues; (e) \textit{\textbf{state}}: recognizes the closed door based on visual interpretation of object state; and (f) \textit{\textbf{order}}: selects the bookshelf by reasoning about relative spatial placement.} 
    \label{fig:teaser}
\end{center}
}]

\begin{abstract}
3D Visual Grounding (3DVG) seeks to locate target objects in 3D scenes using natural language descriptions, enabling downstream applications such as augmented reality and robotics. Existing approaches typically rely on labeled 3D data and predefined categories, limiting scalability to open-world settings. We present \textit{\textbf{\textcolor{BurntOrange}{See}\textcolor{Emerald}{Ground}}}, a zero-shot 3DVG framework that leverages 2D Vision-Language Models (VLMs) to bypass the need for 3D-specific training. To bridge the modality gap, we introduce a hybrid input format that pairs query-aligned rendered views with spatially enriched textual descriptions. Our framework incorporates two core components: a Perspective Adaptation Module that dynamically selects optimal viewpoints based on the query, and a Fusion Alignment Module that integrates visual and spatial signals to enhance localization precision. Extensive evaluations on ScanRefer and Nr3D confirm that \textit{SeeGround} achieves substantial improvements over existing zero-shot baselines -- outperforming them by $7.7\%$ and $7.1\%$, respectively -- and even rivals fully supervised alternatives, demonstrating strong generalization under challenging conditions.
\end{abstract}
\section{Introduction}
\label{sec:intro}

3D Visual Grounding (3DVG) focuses on localizing referred objects within 3D scenes using natural language descriptions. This capability is central to applications in augmented reality~\cite{chen2020scanrefer, liu2023nips, liu2021tmm, liu2024cvpr, wei2024icme, ma2023examination}, vision-language navigation~\cite{chen2022think, huang2022assister, gong2024cognition}, and robotic perception~\cite{chen2023clip2scene, kong2023robo3d, kong2023rethinking, lai2023xvo, li2022coarse3d, zhuang2021perception, tan2024epmf, li2024tfnet, zhuang2024robust, hu2024dhp, kong2025lasermix++, bian2025dynamiccity, kong2025calib3d}. Tackling this task demands both linguistic comprehension and spatial reasoning in cluttered and diverse 3D environments.

Most existing approaches rely on training task-specific models~\cite{jain2022bottom, 3dvista, eda, zhao20213dvg, yuan2021instancerefer, chen2020scanrefer, mcln} using limited, annotation-heavy datasets, which constrains their generalizability. Expanding these models to broader settings is both resource-intensive and impractical~\cite{behley2019semanticKITTI, sun2020waymoOpen, fong2022panoptic-nuScenes}. Recent trends~\cite{zsvg3d, llmgrounder} attempt to mitigate the reliance on 3D supervision by incorporating large language models (LLMs)~\cite{gpt35, openai2023gpt4} to interpret reformatted text queries. However, these strategies often neglect crucial visual attributes -- such as color, texture, perspective, and spatial layout -- that are essential for fine-grained grounding (see \cref{fig:teaser}).

To overcome these limitations, we introduce \textit{\textbf{\textcolor{BurntOrange}{See}\textcolor{Emerald}{Ground}}}, a training-free 3DVG framework that capitalizes on the open-vocabulary capabilities of 2D Vision-Language Models (VLMs) \cite{openai2023gpt4, qwen2-vl, cogvlm2}. These models, pretrained on large-scale image-text corpora, exhibit strong generalization, making them ideal for zero-shot 3DVG \cite{jia2023sceneverse, 3dvista}. Since VLMs are not inherently designed for 3D inputs, we propose a \textbf{cross-modal alignment} mechanism that reformulates 3D scenes into compatible inputs through query-driven renderings and spatially enriched textual descriptions. This strategy enables reasoning over 3D content without additional 3D-specific training~\cite{li2025seeground}.

Our representation combines a rendered 2D image aligned with the query and structured spatial text derived from precomputed object detections. Unlike static multi-view or bird’s-eye projections, our query-guided rendering dynamically captures both local object detail and global context. The spatial text contributes precise semantic and positional cues. To further bridge the gap between language and vision, we incorporate a \textbf{visual prompting} technique that highlights candidate regions, guiding the VLM to resolve ambiguities and attend to relevant image areas.

We validate our approach on two standard benchmarks. On \textit{ScanRefer}\cite{chen2020scanrefer}, SeeGround achieves a $7.7\%$ improvement over prior zero-shot methods, and on \textit{Nr3D}\cite{achlioptas2020referit3d}, it improves by $7.1\%$, narrowing the gap to fully supervised models. Notably, our method remains robust under ambiguous or partial language inputs by relying on visual context to complete the grounding process.

To summarize, our contributions are as follows:
\begin{itemize}
    \item We present \textit{\textbf{\textcolor{BurntOrange}{See}\textcolor{Emerald}{Ground}}}, a training-free method for zero-shot 3DVG, which reformulates 3D scenes into inputs suitable for 2D-VLMs via rendered views and spatial text.
    \item We design a query-guided viewpoint selection strategy to capture both object-specific cues and spatial context.
    \item We propose a visual prompting mechanism to align 2D image features with 3D spatial descriptions, reducing grounding ambiguity in cluttered scenes.
    \item Our approach achieves state-of-the-art zero-shot results on \textit{ScanRefer} and \textit{Nr3D}, demonstrating strong generalization without requiring 3D-specific training.
\end{itemize}
\section{Related Work}

\noindent\textbf{3D Visual Grounding.}  
Supervised 3DVG methods aim to align 3D spatial data with natural language queries, often relying on carefully curated annotations. Early works like ScanRefer~\cite{chen2020scanrefer} and ReferIt3D~\cite{achlioptas2020referit3d} introduced attention-based architectures such as 3DVG-Transformer~\cite{zhao20213dvg} to model these cross-modal correspondences. Subsequent efforts enhanced this foundation through improved fusion techniques: ViewRefer~\cite{viewrefer} incorporates LLM-driven semantics, MVT~\cite{mvt} and LAR~\cite{lar} integrate multi-view geometric reasoning, while SAT~\cite{sat} introduces 2D-guided supervision. Transformer-based designs and weak supervision approaches such as BUTD-DETR~\cite{jain2022bottom}, ConcreteNet~\cite{concretenet}, and WS-3DVG~\cite{ws3dvg} further boost performance. PQ3D~\cite{pq3d} extends grounding to a broader suite of 3D vision-language tasks under a unified framework. Despite their effectiveness, these models depend heavily on dense 3D annotations. Recent zero-shot alternatives -- \eg, LLM-Grounder~\cite{llmgrounder} and ZSVG3D~\cite{zsvg3d} -- remove this supervision requirement but struggle to handle fine-grained visual cues critical for precise localization.

\noindent\textbf{3D Open-Vocabulary Understanding.}  
To generalize beyond fixed taxonomies, recent research explores open-vocabulary 3D understanding by transferring 2D vision-language knowledge into 3D domains~\cite{chen2025ovgaussian, liu2024m3net, chen2023towards, xu2025limoe, lu2025geal}. OpenScene~\cite{peng2023openscene} maps CLIP-derived features into 3D spaces for segmentation tasks, while LeRF~\cite{kerr2023lerf} fuses CLIP with NeRF to capture semantic radiance. Multi-view approaches like OVIR-3D~\cite{ovir3d} and Agent3D-Zero~\cite{agent3d-zero} facilitate instance retrieval and spatial QA. Other techniques -- RegionPLC~\cite{regionplc}, OpenMask3D~\cite{openmask3d}, and OpenIns3D~\cite{openins3d} -- apply 2D cues to supervise 3D perception pipelines. More recently, SAI3D~\cite{sai3d} has incorporated SAM-based segmentation into 3D graph-based reasoning, further validating the strength of 2D supervision for 3D tasks.

\noindent\textbf{MLLMs for 3D Perception.}  
Multimodal large language models (MLLMs) have expanded their utility from 2D grounding to a variety of 3D perception tasks~\cite{kong2023lasermix, liu2023seal, xu2024superflow, xu2025frnet}. Scene-LLM~\cite{fu2023scene-llm} and Uni3DL~\cite{li2023uni3dl} extend MLLMs to 3D captioning and segmentation, while 3D-ViSTA~\cite{3dvista} and ConceptFusion~\cite{conceptfusion2023} align 3D spatial features with language using transformer-based architectures. GLOVER~\cite{Ma2024GLOVER} enables open-vocabulary manipulation tasks, and SceneVerse~\cite{jia2023sceneverse} provides richly annotated 3D environments to support spatial reasoning. RLHF-V~\cite{sun2024interactive} incorporates reinforcement learning to train models for instruction-following in partially observable environments. Our work builds upon this emerging direction by proposing a training-free, zero-shot framework that aligns vision-language models with 3D scenes -- without requiring any 3D-specific fine-tuning or annotations.
\begin{figure*}[t]
    \centering
    \includegraphics[width=\linewidth]{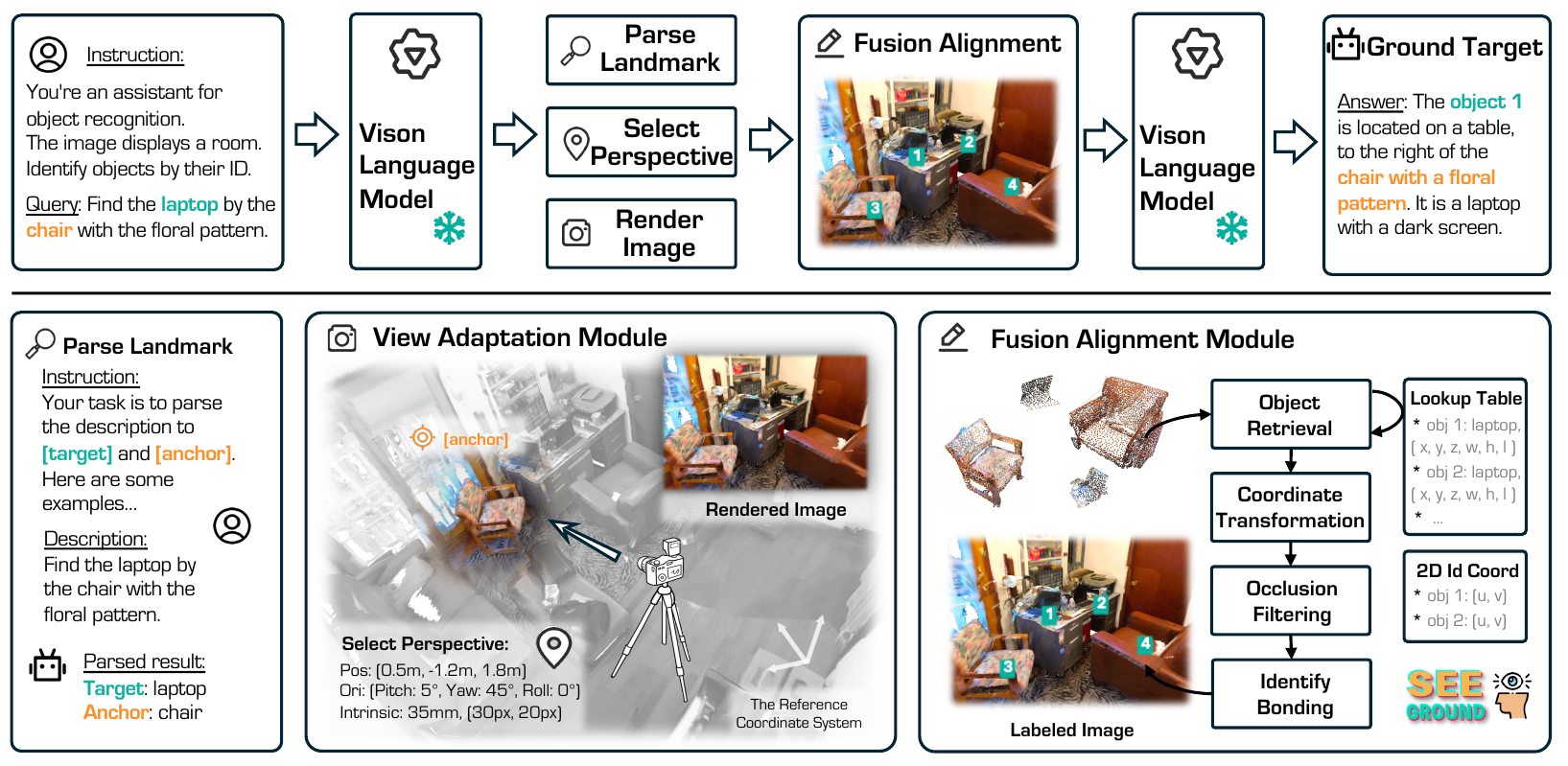}
    \vspace{-0.7cm}
    \caption{Overview of the \textit{\textbf{\textcolor{BurntOrange}{See}\textcolor{Emerald}{Ground}}} framework. A 2D-VLM first interprets the query, identifying the target (\eg, ``laptop'') and an anchor (\eg, \textit{``chair with a floral pattern''}). A dynamic viewpoint is selected based on the anchor’s position to render a query-aligned 2D image. Using the Object Lookup Table ($\mathcal{OLT}$), we retrieve 3D boxes, project visible ones, and apply visual prompts to reduce occlusion. The prompted image, spatial text, and query are fed into the 2D-VLM to localize the target. The predicted ID is then used to retrieve its 3D bounding box from $\mathcal{OLT}$.}
    \label{fig:fig2}
    \vspace{-0.2cm}
\end{figure*}
\begin{figure}
    \centering
    \includegraphics[width=\linewidth]{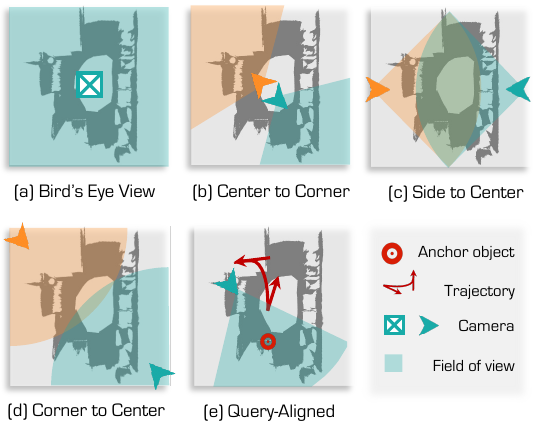}    
    \vspace{-0.6cm}
    \caption{Illustrative example of different perspective selection strategies. Our ``Query-Aligned" method dynamically adapts the viewpoint to match the spatial context of the query, enhancing detail and relevance of visible objects compared to static methods.} 
    \label{fig:view-selection}
    \vspace{-0.2cm}
\end{figure}

\section{Methodology}
\label{sec:3}

\noindent\textbf{Overview.}  
The goal of 3D Visual Grounding (3DVG) is to localize a target object within a 3D scene $\mathcal{S}$ based on a natural language query $\mathsf{Q}$, by predicting its corresponding 3D bounding box:
\[
    \mathbf{bbox} = \mathbf{3DVG}(\mathcal{S}, \mathsf{Q}).
\]
We present a novel 3DVG framework that leverages 2D vision-language models (2D-VLMs) in conjunction with spatially enriched 3D representations. Since conventional 3D data formats are incompatible with the input modalities of 2D-VLMs, we propose a \textbf{hybrid representation} that fuses rendered 2D views with structured 3D spatial descriptions. This allows 2D-VLMs to jointly reason over visual and spatial information without 3D-specific retraining.

Our framework consists of three main components: (1) a multimodal 3D representation module (\cref{sec:3.1}); (2) a Perspective Adaptation Module (\cref{sec:3.2}); and (3) a Fusion Alignment Module (\cref{sec:3.3}). This architecture enables accurate interpretation and localization of objects in complex 3D scenes by fully utilizing the strengths of pretrained 2D-VLMs. The framework overview is illustrated in \cref{fig:fig2}.

\subsection{Multimodal 3D Representation}
\label{sec:3.1}

We leverage 2D vision-language models (2D-VLMs) pretrained on large-scale image-text data to enable open-set understanding of novel objects. However, conventional 3D representations -- such as point clouds~\cite{peng2023openscene, hong2023injecting}, voxels~\cite{li2024is}, and implicit fields~\cite{kerr2023lerf} -- are inherently incompatible with the input format expected by 2D-VLMs. To bridge this gap, we propose a \textbf{hybrid representation} that combines 2D rendered images with text-based 3D spatial descriptions.

\noindent\textbf{Text-based 3D Spatial Descriptions.} 
We begin by detecting all objects in the scene with an open-vocabulary 3D detector:
\[
(\mathbf{bbox}, \mathbf{sem} )_{i=1}^{N} = \mathbf{OVDet}(\mathcal{S}),
\]
where $\mathbf{bbox}$ and $\mathbf{sem}$ denote the 3D bounding box and semantic label of each object, respectively. These outputs are converted into natural language and stored in an object lookup table (OLT) for reuse:
\[
\mathcal{OLT} = \left\{ \left(\mathbf{bbox}, \mathbf{sem} \right) \right\}_{i=1}^{N}.
\]
The $\mathcal{OLT}$ serves as a structured repository of object-level spatial information, supporting efficient reasoning and avoiding redundant computation across multiple queries.

\noindent\textbf{Hybrid 3D Scene Representation.}  
While text descriptions encode layout and semantics, they lack fine-grained visual cues. To complement this, we render a 2D image aligned with the input query:
\[
(\mathbf{I}, \mathcal{T}) = \mathbf{F}(\mathcal{S}, \mathsf{Q}, \mathcal{OLT}),
\]
where $\mathbf{I}$ is the rendered image and $\mathcal{T}$ is the corresponding spatial description text. This pairing enables the 2D-VLM to jointly access visual appearance cues (\eg, color, texture, shape) and accurate 3D spatial semantics, facilitating comprehensive scene understanding.

\begin{table*}[t]
    \centering
    % \caption{Evaluations of 3DVG on \textit{ScanRefer}~\cite{chen2020scanrefer} validation set. Results are reported for \textit{``Unique"} (scenes with a single target object) and \textit{``Multiple"} (scenes with distractors of the same class) subsets, along with overall performance. * indicates results on selected 250 samples.}
    \caption{Results on \textit{ScanRefer}~\cite{chen2020scanrefer} validation set. * denotes evaluation on 250 selected samples.}
    \vspace{-0.2cm}
    \resizebox{\linewidth}{!}{
    \begin{tabular}{r|r|c|c|cc|cc|cc}
        \toprule 
        \multirow{2}{*}{\textbf{Method}} & \multirow{2}{*}{\textbf{Venue}} & \multirow{2}{*}{\textbf{Supervision}} & \multirow{2}{*}{\textbf{Agent}} & \multicolumn{2}{c|}{\textbf{Unique}} & \multicolumn{2}{c|}{\textbf{Multiple}} & \multicolumn{2}{c}{\textbf{Overall}} 
        \\
        & & & & \textbf{Acc@$\mathbf{0.25}$} & \textbf{Acc@$\mathbf{0.5}$} & \textbf{Acc@$\mathbf{0.25}$} & \textbf{Acc@$\mathbf{0.5}$} & \textbf{Acc@$\mathbf{0.25}$} & \textbf{Acc@$\mathbf{0.5}$} 
        \\
        \midrule\midrule
        % ScanRefer \cite{chen2020scanrefer} & ECCV'20 & Fully & - & $65.0$ & $43.3$ & $30.6$ & $19.8$ & $37.3$ & $24.3$
        ScanRefer \cite{chen2020scanrefer} & ECCV'20 & Fully & - & $67.6$ & $46.2$ & $32.1$ & $21.3$ & $39.0$ & $26.1$
        \\
        % TGNN \cite{huang2021text} & AAAI'21 & Fully & - & $64.5$ & $53.0$ & $27.0$ & $21.9$ & $34.3$ & $29.7$ 
        % TGNN \cite{huang2021text} & AAAI'21 & Fully & - & $68.6$ & $56.8$ & $29.8$ & $23.2$ & $37.4$ & $29.7$ 
        % \\
        InstanceRefer \cite{yuan2021instancerefer} & ICCV'21 & Fully & - & $77.5$ & $66.8$ & $31.3$ & $24.8$ & $40.2$ & $32.9$ 
        \\
        % 3DVG-Transformer \cite{zhao20213dvg} & ICCV'21 & Fully & - & $81.9$ & $60.6$ & $39.3$ & $28.4$ & $47.6$ & $34.7$ 
        3DVG-T \cite{zhao20213dvg} & ICCV'21 & Fully & - & $77.2$ & $58.5$ & $38.4$ & $28.7$ & $45.9$ & $34.5$ 
        \\
        BUTD-DETR \cite{jain2022bottom} & ECCV'22 & Fully & - & $84.2$ & $66.3$ & $46.6$ & $35.1$ & $52.2$ & $39.8$
        \\
        EDA \cite{eda} & CVPR'23 & Fully & - & $85.8$ & $68.6$ &  $49.1$  & $37.6$  & $54.6$ &  $42.3$
        \\
        3D-VisTA \cite{3dvista} & ICCV'23 & Fully & - & $81.6$ & $75.1$ & $43.7$ & $39.1$  & $50.6$ &   $45.8$ 
        \\
        G3-LQ \cite{g3lq} & CVPR'24 & Fully & - & $88.6$ & $73.3$ & $50.2$ & $39.7$  & $56.0$ &   $44.7$ 
        \\
        MCLN \cite{mcln} & ECCV'24 & Fully & - & $86.9$ & $72.7$ & $52.0$ & $40.8$  & $57.2$ &   $45.7$ 
        \\
        ConcreteNet \cite{concretenet} & ECCV'24 & Fully & - & $86.4$ & $82.1$ & $42.4$ & $38.4$  & $50.6$ &   $46.5$ 
        \\
        \midrule
        WS-3DVG \cite{ws3dvg}  & ICCV'23 & Weakly & - & - & - & - & -  & $27.4$ &   $22.0$ 
        \\
        \midrule
        LERF \cite{kerr2023lerf} & ICCV'23 & Zero-Shot & CLIP \cite{clip} & - & -  & - & - & $4.8$ & $0.9$ 
        \\
        OpenScene \cite{peng2023openscene} & CVPR'23 & Zero-Shot & CLIP \cite{clip} & $20.1$ & $13.1$ & $11.1$ & $4.4$ & $13.2$ & $6.5$ 
        \\
        LLM-G \cite{llmgrounder} & ICRA'24 & Zero-Shot & GPT-3.5 \cite{gpt35} & - & -  & - & - & $14.3$ & $4.7$ 
        \\
        LLM-G \cite{llmgrounder} & ICRA'24 & Zero-Shot & GPT-4 turbo \cite{openai2023gpt4} & - & -  & - & - & $17.1$ & $5.3$ 
        \\
        ZSVG3D \cite{zsvg3d} & CVPR'24 & Zero-Shot & GPT-4 turbo \cite{openai2023gpt4} & $63.8$ & $58.4$ & $27.7$ & $24.6$ & $36.4$ & $32.7$
        \\
        % ZSVG3D$^\dagger$ \cite{zsvg3d} & CVPR'24 & Zero-Shot & Qwen2-VL-72b \cite{qwen2-vl} & $58.3$ & $53.2$ & \textcolor{red}{todo} & \textcolor{red}{todo} & $40.9$ & $36.8$
        % \\
        VLM-Grounder* \cite{vlmgrounder} & CoRL'24 & Zero-Shot & GPT-4V \cite{openai2023gpt4} & $66.0$ & $29.8$ & $\mathbf{48.3}$ & $\mathbf{33.5}$ & $\mathbf{51.6}$ & $32.8$
        \\
        \textbf{\textcolor{BurntOrange}{See}\textcolor{Emerald}{Ground}} & \textbf{Ours} & Zero-Shot & Qwen2-VL-72b \cite{qwen2-vl} & $\mathbf{75.7}$ & $\mathbf{68.9}$ & $34.0$ & $30.0$ & $44.1$ & $\mathbf{39.4}$
        \\
        \bottomrule
    \end{tabular}}
    \label{tab:tab1}
    \vspace{-0.3cm}
\end{table*}

\subsection{Perspective Adaptation Module}
\label{sec:3.2}
Existing view selection strategies often fail to align with the perspective implied by the query. For instance, LAR~\cite{lar} renders object-centric multi-views but lacks global scene context, while a bird’s-eye view offers comprehensive spatial coverage but omits vertical information, resulting in occlusions and misinterpretations (see \cref{fig:view-selection}(a)). Multi-view or multi-scale approaches~\cite{openins3d} improve coverage (see \cref{fig:view-selection}(b)--(d)), but still rely on static viewpoints. Moreover, 2D-VLMs can misinterpret scenes when the rendered perspective does not reflect the linguistic query. Thus, we introduce a query-driven dynamic rendering strategy that aligns the viewpoint with the query intent, capturing more relevant spatial and visual details (see \cref{fig:view-selection}(e)).

\noindent\textbf{Dynamic Perspective Selection.}  
Given a query $\mathsf{Q}$, the 2D-VLM identifies an anchor object $\boldsymbol{A}$ and a set of candidate targets $\mathcal{O}^{(C)}$ using few-shot prompts $\mathcal{E}^{(E)}$:
\[
  \left( \boldsymbol{A}, \mathcal{O}^{(C)} \right) = \operatorname{VLM} \left( \mathsf{Q}, \mathcal{E}^{(E)} \right).
\]
We place the virtual camera at the scene center, facing the anchor object $\boldsymbol{A}$, and shift it backward and upward to enhance visibility and context. If no anchor can be confidently extracted (\eg, in multi-object or ambiguous queries), we default to a pseudo-anchor located at the centroid of $\mathcal{O}^{(C)}$, and apply the same camera placement strategy.

\noindent\textbf{Query-Aligned Image Rendering.}  
Based on the selected viewpoint, we compute the camera pose using a $\operatorname{look\text{-}at\text{-}view\text{-}transform}$ function, which produces rotation $\mathbf{R}_c$ and translation $\mathbf{T}_c$ with respect to $\boldsymbol{A}$. The rendered image is then obtained as $\mathbf{I} = \operatorname{Render}(\mathcal{S}, \mathbf{R}_c, \mathbf{T}_c)$.
This query-aligned rendering preserves critical visual features while filtering out irrelevant clutter, enabling the 2D-VLM to more accurately localize the referred object (see \cref{fig:view-selection}(e)).

\begin{table}[t]
    \centering
    \caption{Performance on \textit{Nr3D}~\cite{achlioptas2020referit3d}. \textit{Easy}/\textit{Hard}: based on distractor count; \textit{View-Dep.}/\textit{View-Indep.}: based on viewpoint sensitivity.}
    \vspace{-0.2cm}
    \resizebox{\linewidth}{!}{
    \begin{tabular}{r|cccc|c}
        \toprule
        \textbf{Method} & \textbf{Easy} & \textbf{Hard} & \textbf{Dep.} & \textbf{Indep.} & \textbf{Overall} 
        \\
        \midrule\midrule
        \multicolumn{6}{l}{\textbf{Supervision: Fully Supervised}}
        \\
        ReferIt3DNet \cite{achlioptas2020referit3d} & $43.6$ & $27.9$ & $32.5$ & $37.1$ & $35.6$ \\
        TGNN \cite{huang2021text} & $44.2$ & $30.6$ & $35.8$ & $38.0$ & $37.3$ \\
        InstanceRefer \cite{yuan2021instancerefer} & $46.0$ & $31.8$ & $34.5$ & $41.9$ & $38.8$ \\
        3DVG-T \cite{zhao20213dvg} & $48.5$ & $34.8$ & $34.8$ & $43.7$ & $40.8$ \\
        BUTD-DETR \cite{jain2022bottom} & $60.7$ & $48.4$ & $46.0$ & $58.0$ & $54.6$ \\
        MiKASA \cite{Chang2024MiKASAM}  & $69.7$ & $59.4$ & $65.4$ & $64.0$ & $64.4$ \\
        ViL3DRel \cite{Chen2022vil3dref}  & $70.2$ & $57.4$ & $62.0$ & $64.5$ & $64.4$ \\
        \midrule
        \multicolumn{6}{l}{\textbf{Supervision: Weakly Supervised}} 
        \\
        WS-3DVG \cite{ws3dvg} & $27.3$ & $18.0$ & $21.6$ & $22.9$ & $22.5$ \\
        \midrule
        \multicolumn{6}{l}{\textbf{Supervision: Zero-Shot}} 
        \\
        ZSVG3D \cite{zsvg3d} & $46.5$ & $31.7$ & $36.8$ & $40.0$ & $39.0$ \\
        \textbf{\textcolor{BurntOrange}{See}\textcolor{Emerald}{Ground}} & $\mathbf{54.5}$ & $\mathbf{38.3}$ & $\mathbf{42.3}$ & $\mathbf{48.2}$ & $\mathbf{46.1}$ \\
        \bottomrule
    \end{tabular}
    }
  \vspace{-0.3cm}
    \label{tab:tab2}
\end{table}

\subsection{Fusion Alignment Module}
\label{sec:3.3}
While 2D images and spatial descriptions provide complementary information, directly feeding them into a 2D-VLM may fail to associate visual cues with corresponding 3D semantics -- especially in scenes containing similar instances (\eg, multiple chairs) -- which often leads to grounding errors. To address this, we introduce a \textbf{Fusion Alignment Module} that explicitly aligns 2D visual features with spatially grounded object descriptions.

\noindent\textbf{Depth-Aware Visual Prompting.}  
Given the rendered image $\mathbf{I}$, we retrieve the 3D points of each object from the object lookup table $\mathcal{OLT}$ and project them onto the image plane using the camera pose $(\mathbf{R}_c, \mathbf{T}_c)$. To handle occlusions, we compare the depth of each point with the rendered depth map and retain only visible points. For each object $o$, we place a visual prompt $\mathcal{M}_o$ at the center of its visible projection. The prompted image $\mathbf{I}_m$ is generated as:
\[
\mathbf{I}_m = \mathbf{I} \odot \left( 1 - \mathds{1}_{\mathcal{P}_{\mathrm{visible}}(o)} \right) + \mathcal{M}_o \odot \mathds{1}_{\mathcal{P}_{\mathrm{visible}}(o)},
\]
where $\mathds{1}_{\mathcal{P}_{\mathrm{visible}}(o)}$ is an indicator mask for the visible pixels belonging to object $o$.

\noindent\textbf{Object Prediction with 2D-VLM.}  
Finally, given the natural language query $\mathsf{Q}$, the prompted image $\mathbf{I}_m$, and the structured spatial description $\mathcal{T}$, the 2D-VLM predicts the referred object:
\[
\hat{o} = \operatorname{VLM} \left( \mathsf{Q} \, \big| \, \mathbf{I}_m, \mathcal{T} \right).
\]
By enforcing alignment between visual and spatial modalities, this module effectively reduces grounding ambiguity and improves object localization in cluttered scenes.
\section{Experiments}
\label{sec:exp}

\subsection{Experimental Settings}
\label{sec:4.1}

\noindent\textbf{Datasets.}
We evaluate our method on two widely used 3D visual grounding benchmarks. \textbf{ScanRefer}~\cite{chen2020scanrefer} contains 51,500 referring expressions across 800 ScanNet scenes. \textbf{Nr3D}~\cite{achlioptas2020referit3d}, includes 41,503 queries collected through a two-player game. ScanRefer focuses on sparse point cloud grounding, while Nr3D provides dense 3D bounding box annotations, enabling more fine-grained evaluation.

\noindent\textbf{Implementation Details.}
Ablation experiments are conducted on the Nr3D validation split. Images are rendered at $1000{\times}1000$ resolution, excluding the top $0.3$\,m to match closed-room settings. We follow ZSVG3D~\cite{zsvg3d} and use Mask3D~\cite{openmask3d} for consistent object detection.

\subsection{Comparative Study}
On \textbf{ScanRefer}, our method achieves $75.7\%$ / $68.9\%$ at Acc@0.25 / Acc@0.5 on the \textit{``Unique''} split, and $34.0\%$ / $30.0\%$ on the \textit{``Multiple''} split, surpassing all existing zero-shot and weakly supervised baselines~\cite{llmgrounder, zsvg3d, ws3dvg}, and approaching the performance of fully supervised methods~\cite{mcln, concretenet}.
On \textbf{Nr3D}, our model attains an overall accuracy of $46.1\%$, outperforming the previous zero-shot state-of-the-art by $+7.1\%$~\cite{zsvg3d}. It remains robust across different subsets, achieving $54.5\%$ / $38.3\%$ on the \textit{``Easy''} / \textit{``Hard''} splits, and $42.3\%$ / $48.2\%$ on the \textit{``View-Dependent''} / \textit{``View-Independent''} splits, effectively narrowing the gap with fully supervised counterparts~\cite{jain2022bottom}.

\subsection{Ablation Study}
\label{sec:4.4}

\noindent\textbf{Effect of Architecture Design.}  
We begin by evaluating the contribution of each component in the proposed architecture. The results are summarized in \cref{tab:aba}.

\begin{figure}[t]
    \centering
    \includegraphics[width=\linewidth]{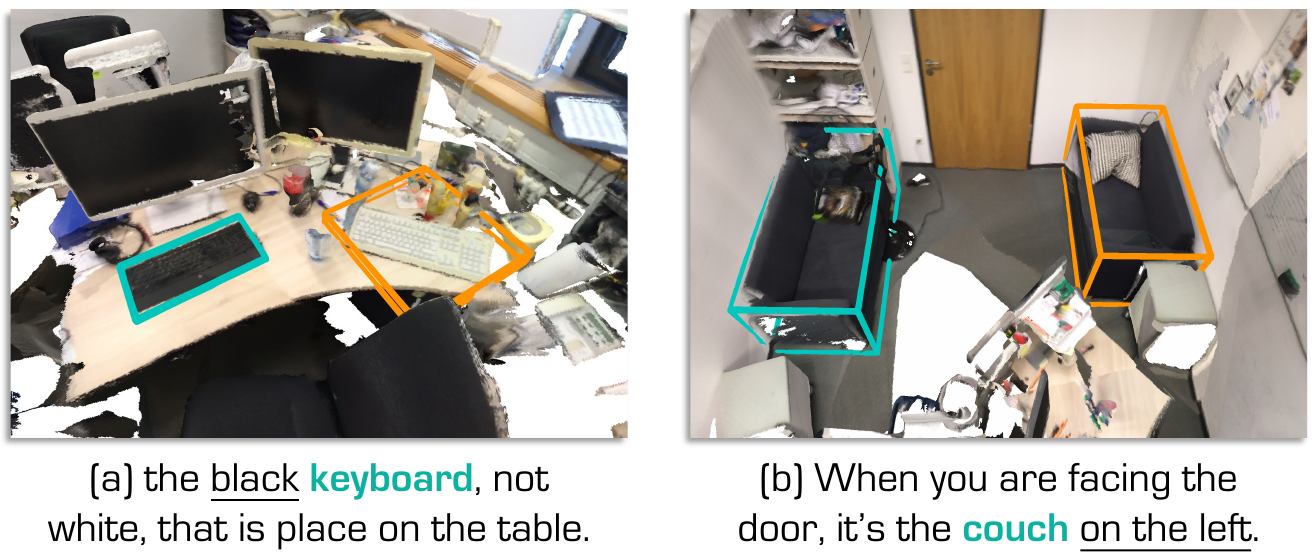}
    \vspace{-0.7cm}
    \caption{\noindent\textbf{Qualitative Results.}
    Rendered scenes with model predictions: correct objects in {\color{Emerald}{\textbf{Green}}}, incorrect in {\color{BurntOrange}{\textbf{Orange}}}. Key \underline{visual cues} (\eg, color, texture, spatial relations) are underlined to illustrate the model's reasoning.}
    \vspace{-0.3cm}
    \label{fig:vis}
\end{figure}

\begin{itemize}
    \item \textit{Layout of the Scene.} Using only 3D coordinates (\textbf{37.7\%}, \cref{tab:aba}(a)) provides coarse object locations but yields low accuracy. Incorporating scene layout (\textbf{39.7\%}, \cref{tab:aba}(b)), via 2D renderings of 3D bounding boxes without texture or color, introduces spatial context that helps the model reason about object size and position.

    \item \textit{Visual Clues.} Integrating object color/texture (\textbf{39.5\%}, \cref{tab:aba}(c)) allows the model to differentiate between visually similar objects, \eg, ``white'' \vs ``black'' (\cref{fig:vis}(a)).

    \item \textit{Fusion Alignment Module.} As shown in \cref{tab:aba}(d), adding our proposed Fusion Alignment Module boosts accuracy to \textbf{43.3\%} by aligning rendered images with spatial text, enabling the model to ground targets in cluttered scenes.

    \item \textit{Perspective Adaptation Module.} Incorporating the Perspective Adaptation Module (\textbf{45.0\%}, \cref{tab:aba}(e)) improves grounding accuracy by aligning the viewpoint with the spatial context implied by the query (\cref{fig:vis}(b)). This helps resolve ambiguities and enhances spatial reasoning.

    \item \textit{Full Configuration.} The highest accuracy (\textbf{46.1\%}) is achieved with the complete configuration (\cref{tab:aba}(f)), validating the effectiveness of \textsc{SeeGround} and the synergistic benefit of combining all components.
\end{itemize}

\begin{table}[t]
    \centering
    \caption{\textbf{Ablation study}. \textit{``3D Pos."}: Object coordinates; \textit{``Layout"}: Scene structure; \textit{``Texture"}: Color/texture; \textit{``FAM"}: Fusion Alignment Module; \textit{``PAM"}: Perspective Adaptation Module.}
    \vspace{-0.2cm}
    \resizebox{\linewidth}{!}{
    \begin{tabular}{c|ccccc|c}
    \toprule
    \textbf{\#} & \textbf{3D Pos.} & \textbf{Layout} & \textbf{Texture} & \textbf{FAM} & \textbf{PAM} & \textbf{Overall} 
    \\
    \midrule \midrule
    (a) & \textcolor{Emerald}{\cmark} & \textcolor{BurntOrange}{\xmark} & \textcolor{BurntOrange}{\xmark} & \textcolor{BurntOrange}{\xmark} & \textcolor{BurntOrange}{\xmark} & $37.7$ 
    \\
    (b) & \textcolor{Emerald}{\cmark} & \textcolor{Emerald}{\cmark} & \textcolor{BurntOrange}{\xmark} & \textcolor{BurntOrange}{\xmark} & \textcolor{BurntOrange}{\xmark} & $39.7$ 
    \\
    (c) & \textcolor{Emerald}{\cmark} & \textcolor{BurntOrange}{\xmark} & \textcolor{Emerald}{\cmark} & \textcolor{BurntOrange}{\xmark} & \textcolor{BurntOrange}{\xmark} & $39.5$ 
    \\
    (d) & \textcolor{Emerald}{\cmark} & \textcolor{Emerald}{\cmark} & \textcolor{Emerald}{\cmark} & \textcolor{Emerald}{\cmark} & \textcolor{BurntOrange}{\xmark} & $43.3$ 
    \\
    (e) & \textcolor{BurntOrange}{\xmark} & \textcolor{Emerald}{\cmark} & \textcolor{Emerald}{\cmark} & \textcolor{Emerald}{\cmark} & \textcolor{Emerald}{\cmark} & $45.0$ 
    \\\midrule
    (f) & \textcolor{Emerald}{\cmark} & \textcolor{Emerald}{\cmark} & \textcolor{Emerald}{\cmark} & \textcolor{Emerald}{\cmark} & \textcolor{Emerald}{\cmark} & $\mathbf{46.1}$
    \\
    \bottomrule
\end{tabular}
\vspace{-0.5cm}
}
\label{tab:aba}
\end{table}

\begin{figure}[t]
    \centering
    \begin{subfigure}[h]{0.48\linewidth}
        \centering
        \includegraphics[width=\linewidth]{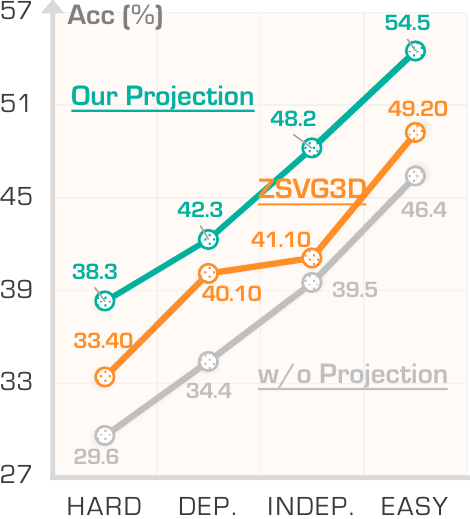}
        \caption{Projection Method}
        \label{fig:compare_projection}
    \end{subfigure}~~
    \begin{subfigure}[h]{0.48\linewidth}
        \centering
        \includegraphics[width=\linewidth]{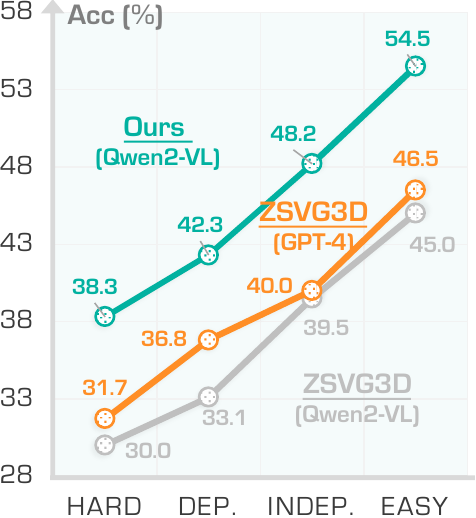}
        \caption{Language Agent}
        \label{fig:compare_agent}
    \end{subfigure}
    \vspace{-0.25cm}
    \caption{
    \noindent\textbf{Ablation study} on (a) different projection strategies (ours \vs ZSVG3D~\cite{zsvg3d}), and (b) different language agents (GPT-4~\cite{openai2023gpt4} \vs Qwen2-VL~\cite{qwen2-vl}).}
    \vspace{-0.2cm}
    \label{fig:compare}
\end{figure}

\begin{figure}[t]
    \centering
    \includegraphics[width=\linewidth]{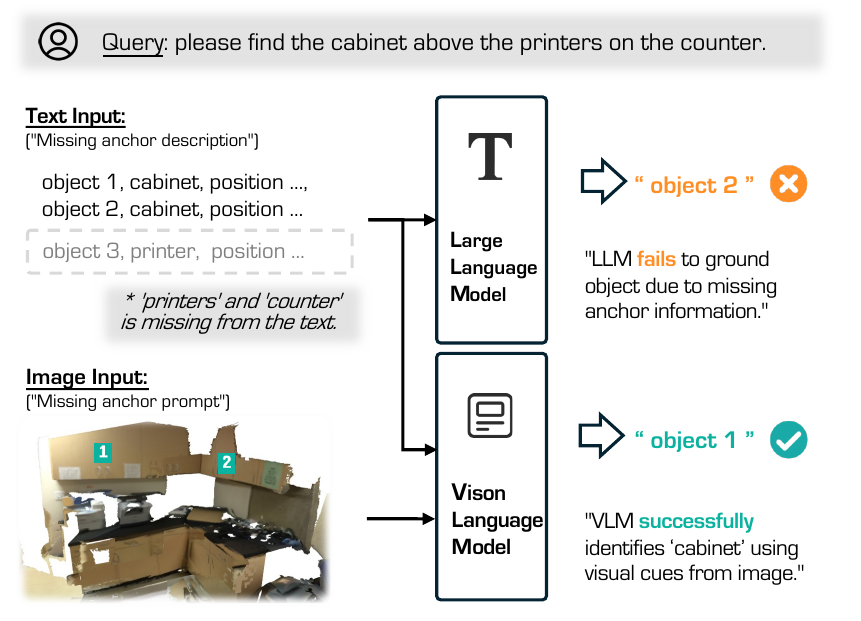}    
    \vspace{-0.2cm}
    \caption{Robustness example: our method correctly identifies the \textit{cabinet} despite missing key textual cues (\eg, \textit{printers}, \textit{counter}) by leveraging visual context, outperforming prior methods that rely more on explicit text.}
    \label{fig:robust}
 \vspace{-0.2cm}
\end{figure}

\noindent\textbf{Ours \vs Prior Art.}  
ZSVG3D~\cite{zsvg3d} infers spatial relations by projecting object centers and applying predefined heuristics, but lacks flexibility, omits visual context, and fails under imperfect detections (\cref{fig:robust}). As shown in \cref{fig:compare_projection}, its VLM-based variant renders only target and anchor centers without background. In contrast, our method produces full-scene renderings, enabling reasoning over undetected or ambiguous objects using surrounding visual cues.

\noindent\textbf{Qwen2-VL \vs GPT-4.}  
To promote accessibility and reproducibility, we adopt the open-source Qwen2-VL~\cite{qwen2-vl} as the agent. For fair comparison, we re-evaluate ZSVG3D using Qwen2-VL in place of GPT-4~\cite{openai2023gpt4} (\cref{fig:compare_agent}). Our method consistently outperforms ZSVG3D under the same VLM, confirming the effectiveness of our strategy, independent of the underlying language model.

\begin{table}[t]
    \centering
    \small
    \caption{
    % Performance comparison of different perspective selection strategies. Our method results in consistently higher accuracy across all difficulty levels on Nr3D \cite{achlioptas2020referit3d} validation set.
    \noindent\textbf{Comparison of Perspective Selection Strategies.}  We compare different viewpoint selection strategies on the \textit{Nr3D}~\cite{achlioptas2020referit3d} validation set. Our method consistently achieves higher accuracy across all difficulty levels, demonstrating the effectiveness of query-aligned dynamic rendering for 3D grounding.
    }
    \vspace{-0.2cm}
    \resizebox{\linewidth}{!}{%
    \begin{tabular}{c|cccc|c}
        \toprule
        \textbf{Type} & \textbf{Easy} & \textbf{Hard} & \textbf{Dep.} & \textbf{Indep.} & \textbf{Overall} 
        \\
        \midrule\midrule
        Center2Corner  & $49.5$ & $31.4$ & $35.1$ & $42.9$  & $40.2$ \\
        Edege2Center   & $51.0$ & $32.7$ & $36.6$ & $44.2$  & $41.5$ \\
        Corner2Center  & $49.8$  & $33.4$  & $35.5$ & $44.5$   & $41.3$ \\
        Bird's Eye View   & $53.4$ & $33.9$ & $36.9$ & $46.8$  & $43.3$ \\
        Query-aligned & $\mathbf{54.5}$ & $\mathbf{38.3}$ & $\mathbf{42.3}$ & $\mathbf{48.2}$ &  $\mathbf{46.1}$ \\
        \bottomrule
    \end{tabular}}
    \vspace{-0.2cm}
    \label{tab:view-selection}
\end{table}

\noindent\textbf{Effect of View Selection Strategy.} 
\cref{tab:view-selection} shows the impact of different viewpoint strategies. Our query-driven approach outperforms static baselines. Fixed methods (\textit{Center2Corner}, \textit{Edge2Center}, \textit{Corner2Center}) lack adaptability, while BEV, though global, misses key spatial cues like orientation and height. In contrast, our dynamic strategy achieves consistent gains, notably on \textit{Hard} ($+4.4\%$) and \textit{View-Dependent} ($+5.7\%$) queries.

\noindent\textbf{Robustness Evaluation with Incomplete Textual Descriptions.}  
\cref{fig:robust} shows our model's robustness under incomplete queries, where anchor objects are omitted to simulate detection failures. While LLM-based methods degrade significantly without anchor cues, our approach successfully leverages visual context to maintain accurate grounding. These results underscore the importance of integrating visual and textual signals for robust 3D understanding.

\begin{table}[t]
    \centering
    % \vspace{-0.1cm}
    \caption{Performance comparison of different 3D detectors on the ScanRefer~\cite{chen2020scanrefer} validation set. Accuracy (Acc.) is reported for each method paired with different 3D detectors.}
    \vspace{-0.15cm}
    \setlength{\tabcolsep}{15pt}
    \begin{tabular}{l|c|c}
        \toprule
        \textbf{Method} & \textbf{3D Detector} & \textbf{Acc.} \\
        \midrule \midrule
        \multirow{2}{*}{ZSVG3D~\cite{zsvg3d}} & Mask3D~\cite{openmask3d} & $36.4$ 
        \\
        & OVIR-3D & $19.3$ 
        \\
        \midrule
        \multirow{3}{*}{ \textbf{\textcolor{BurntOrange}{See}\textcolor{Emerald}{Ground}}} & Ground Truth & $59.5$ 
        \\
        & Mask3D~\cite{Schult23mask3d}& $44.1$ 
        \\
        & OVIR-3D~\cite{ovir3d} & $30.7$ 
        \\
        \bottomrule
    \end{tabular}
    \vspace{-0.2cm}
    \label{tab:3d_comparison}
\end{table}

\noindent\textbf{Results on Different Detectors.} \cref{tab:3d_comparison} compares the performance of different 3D detectors. With Mask3D, our method achieves $44.1\%$, significantly surpassing ZSVG3D ($36.4\%$). Using OVIR-3D, our performance remains higher ($30.7\%$ vs. $19.3\%$). When provided with ground-truth (GT) boxes, our method reaches $59.5\%$, revealing a clear performance upper bound.

\begin{figure}[t]
    \centering
    \begin{subfigure}[h]{0.45\linewidth}
        \centering
        \includegraphics[width=0.95\linewidth]{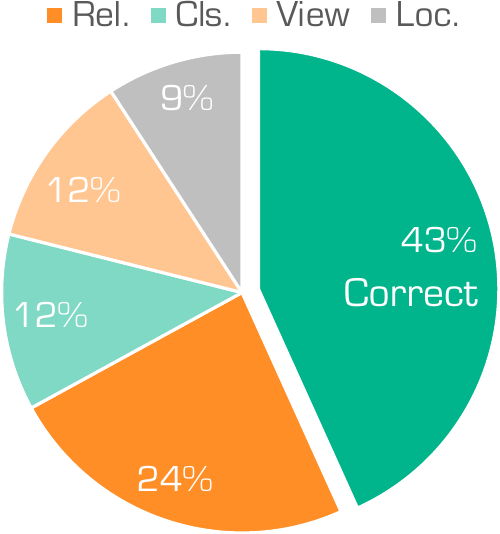}
        \caption{Text-Only Method}
    \end{subfigure}~~~~~~
    \begin{subfigure}[h]{0.45\linewidth}
        \centering
        \includegraphics[width=0.95\linewidth]{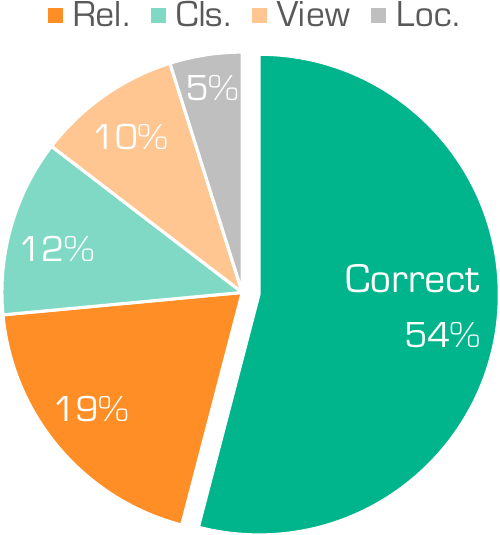}
        \caption{Our Method}
    \end{subfigure}
    \vspace{-0.2cm}
    \caption{Error distributions for the Text-Only method (a) and ours (b), categorized as: \textbf{Rel.} (spatial misinterpretation, \eg, ``next to''), \textbf{Cls.} (category mismatch), \textbf{View} (viewpoint misunderstanding), and \textbf{Loc.} (inaccurate target localization).}
    \label{fig:error_source}
    \vspace{-0.2cm}
\end{figure}

\noindent\textbf{Type-Wise Error Analysis.}  
We analyze 185 randomly sampled cases from 10 scenes to identify common failure modes (\cref{fig:error_source}). Reductions in localization and classification errors demonstrate the benefit of visual input for spatial understanding. However, spatial relation errors remain frequent ($19\%$), suggesting limitations in fine-grained reasoning that could be addressed by dedicated spatial modules.

Our current viewpoint selection also struggles with complex egocentric references (\eg, \textit{``when the window is on the left''}, \textit{``upon entering from the door''}). In addition, limited rendering quality -- due to the use of raw dataset point clouds -- hampers object discrimination. Future work may incorporate high-fidelity rendering to enhance visual clarity in cluttered scenes.
\section{Conclusion}
\label{sec:conclusion}

In this paper, we proposed \textit{\textbf{\textcolor{BurntOrange}{See}\textcolor{Emerald}{Ground}}}, a zero-shot 3D visual grounding framework that bridges 3D data and 2D vision-language models via query-aligned renderings and spatial descriptions. Our Perspective Adaptation Module selects viewpoints dynamically, while the Fusion Alignment Module aligns visual and spatial cues for robust grounding. Experiments on two benchmarks show that our method outperforms zero-shot baselines.

% References
{
    \small
    \bibliographystyle{ieeetr}
    \bibliography{main}

\begin{thebibliography}{10}

\bibitem{chen2020scanrefer}
D.~Z. Chen {\em et~al.}, ``Scanrefer: 3d object localization in rgb-d scans using natural language,'' in {\em ECCV}, pp.~202--221, 2020.

\bibitem{liu2023nips}
Z.~Liu {\em et~al.}, ``Raydf: neural ray-surface distance fields with multi-view consistency,'' {\em arXiv preprint arXiv:2310.19629}, 2023.

\bibitem{liu2021tmm}
Z.~Liu {\em et~al.}, ``Deep view synthesis via self-consistent generative network,'' {\em IEEE Transactions on Multimedia}, vol.~24, pp.~451--465, 2021.

\bibitem{liu2024cvpr}
Z.~Liu {\em et~al.}, ``Unleashing the potential of multi-modal foundation models and video diffusion for 4d dynamic physical scene simulation,'' {\em arXiv preprint arXiv:2411.14423}, 2024.

\bibitem{wei2024icme}
X.~Wei {\em et~al.}, ``Sir: Multi-view inverse rendering with decomposable shadow for indoor scenes,'' {\em arXiv preprint arXiv:2402.06136}, 2024.

\bibitem{ma2023examination}
T.~Ma {\em et~al.}, ``An examination of the compositionality of large generative vision-language models,'' {\em arXiv preprint arXiv:2308.10509}, 2023.

\bibitem{chen2022think}
S.~Chen {\em et~al.}, ``Think global, act local: Dual-scale graph transformer for vision-and-language navigation,'' in {\em CVPR}, pp.~16537--16547, 2022.

\bibitem{huang2022assister}
Z.~Huang {\em et~al.}, ``Assister: Assistive navigation via conditional instruction generation,'' in {\em ECCV}, pp.~271--289, 2022.

\bibitem{gong2024cognition}
Z.~Gong {\em et~al.}, ``From cognition to precognition: A future-aware framework for social navigation,'' {\em arXiv preprint arXiv:2409.13244}, 2024.

\bibitem{chen2023clip2scene}
R.~Chen {\em et~al.}, ``Clip2scene: Towards label-efficient 3d scene understanding by clip,'' in {\em CVPR}, pp.~7020--7030, 2023.

\bibitem{kong2023robo3d}
L.~Kong {\em et~al.}, ``Robo3d: Towards robust and reliable 3d perception against corruptions,'' in {\em ICCV}, pp.~19994--20006, 2023.

\bibitem{kong2023rethinking}
L.~Kong {\em et~al.}, ``Rethinking range view representation for lidar segmentation,'' in {\em ICCV}, pp.~228--240, 2023.

\bibitem{lai2023xvo}
L.~Lai {\em et~al.}, ``Xvo: Generalized visual odometry via cross-modal self-training,'' in {\em ICCV}, pp.~10094--10105, 2023.

\bibitem{li2022coarse3d}
R.~Li {\em et~al.}, ``Coarse3d: Class-prototypes for contrastive learning in weakly-supervised 3d point cloud segmentation,'' {\em arXiv preprint arXiv:2210.01784}, 2022.

\bibitem{zhuang2021perception}
Z.~Zhuang {\em et~al.}, ``Perception-aware multi-sensor fusion for 3d lidar semantic segmentation,'' in {\em ICCV}, pp.~16280--16290, 2021.

\bibitem{tan2024epmf}
M.~Tan {\em et~al.}, ``Epmf: Efficient perception-aware multi-sensor fusion for 3d semantic segmentation,'' {\em TPAMI}, vol.~46, no.~12, pp.~8258--8273, 2024.

\bibitem{li2024tfnet}
R.~Li {\em et~al.}, ``Tfnet: Exploiting temporal cues for fast and accurate lidar semantic segmentation,'' in {\em CVPR}, pp.~4547--4556, 2024.

\bibitem{zhuang2024robust}
Z.~Zhuang {\em et~al.}, ``Robust 3d semantic occupancy prediction with calibration-free spatial transformation,'' {\em arXiv preprint arXiv:2411.12177}, 2024.

\bibitem{hu2024dhp}
T.~Hu {\em et~al.}, ``Dhp-mapping: A dense panoptic mapping system with hierarchical world representation and label optimization techniques,'' in {\em IROS}, pp.~1101--1107, 2024.

\bibitem{kong2025lasermix++}
L.~Kong {\em et~al.}, ``Multi-modal data-efficient 3d scene understanding for autonomous drivin,'' {\em TPAMI}, vol.~47, no.~5, pp.~3748--3765, 2025.

\bibitem{bian2025dynamiccity}
H.~Bian {\em et~al.}, ``Dynamiccity: Large-scale 4d occupancy generation from dynamic scenes,'' {\em arXiv preprint arXiv:2410.18084}, 2024.

\bibitem{kong2025calib3d}
L.~Kong {\em et~al.}, ``Calib3d: Calibrating model preferences for reliable 3d scene understanding,'' in {\em WACV}, pp.~1965--1978, 2025.

\bibitem{jain2022bottom}
A.~Jain {\em et~al.}, ``Bottom up top down detection transformers for language grounding in images and point clouds,'' in {\em ECCV}, pp.~417--433, 2022.

\bibitem{3dvista}
Z.~Zhu {\em et~al.}, ``3d-vista: Pre-trained transformer for 3d vision and text alignment,'' in {\em ICCV}, pp.~2911--2921, 2023.

\bibitem{eda}
Y.~Wu {\em et~al.}, ``Eda: Explicit text-decoupling and dense alignment for 3d visual grounding,'' in {\em CVPR}, pp.~19231--19242, 2023.

\bibitem{zhao20213dvg}
L.~Zhao {\em et~al.}, ``3dvg-transformer: Relation modeling for visual grounding on point clouds,'' in {\em ICCV}, pp.~2928--2937, 2021.

\bibitem{yuan2021instancerefer}
Z.~Yuan {\em et~al.}, ``Instancerefer: Cooperative holistic understanding for visual grounding on point clouds through instance multi-level contextual referring,'' in {\em ICCV}, pp.~1791--1800, 2021.

\bibitem{mcln}
Z.~Qian {\em et~al.}, ``Multi-branch collaborative learning network for 3d visual grounding,'' in {\em ECCV}, pp.~381--398, 2025.

\bibitem{behley2019semanticKITTI}
J.~Behley {\em et~al.}, ``Semantickitti: A dataset for semantic scene understanding of lidar sequences,'' in {\em ICCV}, pp.~9297--9307, 2019.

\bibitem{sun2020waymoOpen}
P.~Sun {\em et~al.}, ``Scalability in perception for autonomous driving: Waymo open dataset,'' in {\em CVPR}, pp.~2446--2454, 2020.

\bibitem{fong2022panoptic-nuScenes}
W.~K. Fong {\em et~al.}, ``Panoptic nuscenes: A large-scale benchmark for lidar panoptic segmentation and tracking,'' {\em RA-L}, vol.~7, pp.~3795--3802, 2022.

\bibitem{zsvg3d}
Z.~Yuan {\em et~al.}, ``Visual programming for zero-shot open-vocabulary 3d visual grounding,'' in {\em CVPR}, pp.~20623--20633, 2024.

\bibitem{llmgrounder}
J.~Yang {\em et~al.}, ``Llm-grounder: Open-vocabulary 3d visual grounding with large language model as an agent,'' in {\em ICRA}, pp.~7694--7701, 2024.

\bibitem{gpt35}
L.~Ouyang {\em et~al.}, ``Training language models to follow instructions with human feedback,'' {\em NeurIPS}, vol.~35, pp.~27730--27744, 2022.

\bibitem{openai2023gpt4}
OpenAI, ``Gpt-4 technical report,'' {\em arXiv preprint arXiv:2303.08774}, 2023.

\bibitem{qwen2-vl}
P.~Wang {\em et~al.}, ``Qwen2-vl: Enhancing vision-language model's perception of the world at any resolution,'' {\em arXiv preprint arXiv:2409.12191}, 2024.

\bibitem{cogvlm2}
W.~Hong {\em et~al.}, ``Cogvlm2: Visual language models for image and video understanding,'' {\em arXiv preprint arXiv:2408.16500}, 2024.

\bibitem{jia2023sceneverse}
B.~Jia {\em et~al.}, ``Sceneverse: Scaling 3d vision-language learning for grounded scene understanding,'' in {\em ECCV}, pp.~289--310, 2025.

\bibitem{li2025seeground}
R.~Li {\em et~al.}, ``Seeground: See and ground for zero-shot open-vocabulary 3d visual grounding,'' {\em arXiv preprint arXiv:2412.04383}, 2024.

\bibitem{achlioptas2020referit3d}
P.~Achlioptas {\em et~al.}, ``Referit3d: Neural listeners for fine-grained 3d object identification in real-world scenes,'' in {\em ECCV}, pp.~422--440, 2020.

\bibitem{viewrefer}
Z.~Guo {\em et~al.}, ``Viewrefer: Grasp the multi-view knowledge for 3d visual grounding,'' in {\em ICCV}, pp.~15372--15383, 2023.

\bibitem{mvt}
S.~Huang {\em et~al.}, ``Multi-view transformer for 3d visual grounding,'' in {\em CVPR}, pp.~15524--15533, 2022.

\bibitem{lar}
E.~M. Bakr {\em et~al.}, ``Look around and refer: 2d synthetic semantics knowledge distillation for 3d visual grounding,'' in {\em NeurIPS}, vol.~35, pp.~37146--37158, 2022.

\bibitem{sat}
Z.~Yang {\em et~al.}, ``Sat: 2d semantics assisted training for 3d visual grounding,'' in {\em ICCV}, pp.~1856--1866, 2021.

\bibitem{concretenet}
O.~Unal {\em et~al.}, ``Four ways to improve verbo-visual fusion for dense 3d visual grounding,'' in {\em ECCV}, pp.~196--213, 2025.

\bibitem{ws3dvg}
Z.~Wang {\em et~al.}, ``Distilling coarse-to-fine semantic matching knowledge for weakly supervised 3d visual grounding,'' in {\em ICCV}, pp.~2662--2671, 2023.

\bibitem{pq3d}
Z.~Zhu {\em et~al.}, ``Unifying 3d vision-language understanding via promptable queries,'' in {\em ECCV}, pp.~188--206, 2024.

\bibitem{chen2025ovgaussian}
R.~Chen {\em et~al.}, ``Ovgaussian: Generalizable 3d gaussian segmentation with open vocabularies,'' {\em arXiv preprint arXiv:2501.00326}, 2025.

\bibitem{liu2024m3net}
Y.~Liu {\em et~al.}, ``Multi-space alignments towards universal lidar segmentation,'' in {\em CVPR}, pp.~14648--14661, 2024.

\bibitem{chen2023towards}
R.~Chen {\em et~al.}, ``Towards label-free scene understanding by vision foundation models,'' in {\em NeurIPS}, pp.~75896--75910, 2023.

\bibitem{xu2025limoe}
X.~Xu {\em et~al.}, ``Limoe: Mixture of lidar representation learners from automotive scenes,'' {\em arXiv preprint arXiv:2501.04004}, 2025.

\bibitem{lu2025geal}
D.~Lu {\em et~al.}, ``Geal: Generalizable 3d affordance learning with cross-modal consistency,'' {\em arXiv preprint arXiv:2412.09511}, 2025.

\bibitem{peng2023openscene}
S.~Peng {\em et~al.}, ``Openscene: 3d scene understanding with open vocabularies,'' in {\em CVPR}, pp.~815--824, 2023.

\bibitem{kerr2023lerf}
J.~Kerr {\em et~al.}, ``Lerf: Language embedded radiance fields,'' in {\em ICCV}, pp.~19729--19739, 2023.

\bibitem{ovir3d}
S.~Lu {\em et~al.}, ``Ovir-3d: Open-vocabulary 3d instance retrieval without training on 3d data,'' in {\em CoRL}, pp.~1610--1620, 2023.

\bibitem{agent3d-zero}
S.~Zhang {\em et~al.}, ``Agent3d-zero: An agent for zero-shot 3d understanding,'' in {\em arXiv preprint arXiv:2403.11835}, 2024.

\bibitem{regionplc}
J.~Yang {\em et~al.}, ``Regionplc: Regional point-language contrastive learning for open-world 3d scene understanding,'' in {\em CVPR}, pp.~19823--19832, 2024.

\bibitem{openmask3d}
A.~Takmaz {\em et~al.}, ``Openmask3d: Open-vocabulary 3d instance segmentation,'' {\em arXiv preprint arXiv:2306.13631}, 2023.

\bibitem{openins3d}
Z.~Huang {\em et~al.}, ``Openins3d: Snap and lookup for 3d open-vocabulary instance segmentation,'' in {\em ECCV}, pp.~169--185, 2025.

\bibitem{sai3d}
Y.~Yin {\em et~al.}, ``Sai3d: Segment any instance in 3d scenes,'' in {\em CVPR}, pp.~3292--3302, 2024.

\bibitem{kong2023lasermix}
L.~Kong {\em et~al.}, ``Lasermix for semi-supervised lidar semantic segmentation,'' in {\em CVPR}, pp.~21705--21715, 2023.

\bibitem{liu2023seal}
Y.~Liu {\em et~al.}, ``Segment any point cloud sequences by distilling vision foundation models,'' in {\em NeurIPS}, vol.~36, pp.~37193--37229, 2023.

\bibitem{xu2024superflow}
X.~Xu {\em et~al.}, ``4d contrastive superflows are dense 3d representation learners,'' in {\em ECCV}, pp.~58--80, 2024.

\bibitem{xu2025frnet}
X.~Xu {\em et~al.}, ``Frnet: Frustum-range networks for scalable lidar segmentation,'' {\em TIP}, vol.~34, pp.~2173--2186, 2025.

\bibitem{fu2023scene-llm}
R.~Fu {\em et~al.}, ``Scene-llm: Extending language model for 3d visual understanding and reasoning,'' {\em arXiv preprint arXiv:2403.11401}, 2024.

\bibitem{li2023uni3dl}
X.~Li {\em et~al.}, ``Uni3dl: A unified model for 3d and language understanding,'' {\em arXiv preprint arXiv:2312.03026}, 2023.

\bibitem{conceptfusion2023}
K.~M. Jatavallabhula {\em et~al.}, ``Conceptfusion: Open-set multimodal 3d mapping,'' {\em Robotics: Science and Systems}, 2023.

\bibitem{Ma2024GLOVER}
T.~Ma {\em et~al.}, ``Glover: Generalizable open-vocabulary affordance reasoning for task-oriented grasping,'' {\em arXiv preprint arXiv:2411.12286}, 2024.

\bibitem{sun2024interactive}
L.~Sun {\em et~al.}, ``Interactive planning using large language models for partially observable robotic tasks,'' in {\em ICRA}, pp.~14054--14061, 2024.

\bibitem{hong2023injecting}
Y.~Hong {\em et~al.}, ``3d-llm: Injecting the 3d world into large language models,'' in {\em NeurIPS}, vol.~36, pp.~20482--20494, 2023.

\bibitem{li2024is}
Y.~Li {\em et~al.}, ``Is your lidar placement optimized for 3d scene understanding?,'' in {\em NeurIPS}, vol.~37, pp.~34980--35017, 2024.

\bibitem{g3lq}
Y.~Wang {\em et~al.}, ``G3-lq: Marrying hyperbolic alignment with explicit semantic-geometric modeling for 3d visual grounding,'' in {\em CVPR}, pp.~13917--13926, 2024.

\bibitem{clip}
A.~Radford {\em et~al.}, ``Learning transferable visual models from natural language supervision,'' in {\em ICML}, pp.~8748--8763, 2021.

\bibitem{vlmgrounder}
R.~Xu {\em et~al.}, ``Vlm-grounder: A vlm agent for zero-shot 3d visual grounding,'' {\em arXiv preprint arXiv:2410.13860}, 2024.

\bibitem{huang2021text}
P.-H. Huang {\em et~al.}, ``Text-guided graph neural networks for referring 3d instance segmentation,'' in {\em AAAI}, vol.~35, pp.~1610--1618, 2021.

\bibitem{Chang2024MiKASAM}
C.-P. Chang {\em et~al.}, ``Mikasa: Multi-key-anchor \& scene-aware transformer for 3d visual grounding,'' in {\em CVPR}, pp.~14131--14140, 2024.

\bibitem{Chen2022vil3dref}
S.~Chen {\em et~al.}, ``Language conditioned spatial relation reasoning for 3d object grounding,'' in {\em NeurIPS}, 2022.

\bibitem{Schult23mask3d}
J.~Schult {\em et~al.}, ``Mask3d: Mask transformer for 3d semantic instance segmentation,'' in {\em ICRA}, pp.~8216--8223, 2023.

\end{thebibliography}
}

\end{document}